\documentclass{article}
\usepackage{spconf,amsmath,graphicx}

\title{Source Camera Identification Based On Content-Adaptive Fusion Network}
\name{Pengpeng Yang, Wei Zhao, Rongrong Ni, and Yao Zhao\thanks{This work was supported in part by National NSF of China (61332012, 61272355, 61672090), Fundamental Research Funds for the Central Universities (2015JBZ002), the PAPD, the CICAEET. We greatly ackonwledge the support of NVIDIA Corporation with the donation of the Tesla K40 GPU used for this research.}}
\address{Institute of Information Science, \\
	\& Beijing Key Laboratory of Advanced Information Science and Network Technology, \\
		Beijing Jiaotong University, Beijing 100044, China\\
		rrni@bjtu.edu.cn, yzhao@bjtu.edu.cn\\}
\begin{document}
%
\maketitle
\begin{abstract}
Source camera identification is still a hard task in forensics community, especially for the case of the small query image size. In this paper, we propose a solution to identify the source camera of the small-size images: content-adaptive fusion network. In order to learn better feature representation from the input data, content-adaptive convolutional neural networks(CA-CNN) are constructed. We add a convolutional layer in preprocessing stage. Moreover, with the purpose of capturing more comprehensive information, we parallel three CA-CNNs: CA3-CNN, CA5-CNN, CA7-CNN to get the content-adaptive fusion network. The difference of three CA-CNNs lies in the convolutional kernel size of pre-processing layer. The experimental results show that the proposed method is practicable and satisfactory.
\end{abstract}
\section{INTRODUCTION}
\label{sec:intro}
\begin{figure*}[!tp]
	\centering
	\includegraphics[width=14cm,height=3cm]{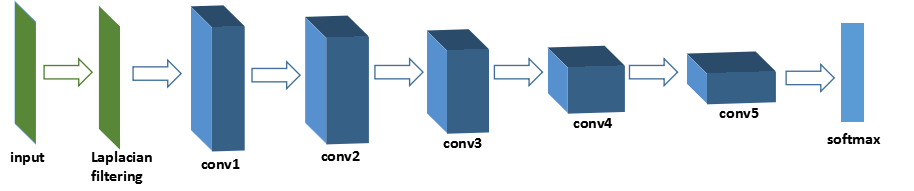}
	\caption{the architecture of LCNN and CA-CNN}
\end{figure*}

With the development of science and technology, image acquisition devices are becoming more and more abundant. At the same time, image editing tools are becoming common and anyone can easily modify the images. So multimedia forensics are needed to prevent malicious tampering to the images for illegal benefits. One of the import topics in multimedia forensics is source camera identification.  

A series of operations inside the camera would be performed when we capture a digital image. These processes could bring some inherent traces left in the image, such as lens aberration \cite{1,2}, defective pixels \cite{3,4}, CFA interpolation artifacts \cite{5}, JPEG compression \cite{6,7}, image quality evaluation index and high order statistics in wavelet domain \cite{8,9} or Sesor Pattern Nosie(SPN) \cite{10,11,12,13,14,15}, which is the key to finding the source camera of an image. Sensor Pattern Noise(SPN) generated by digital cameras has drawn more attention. The SPN arises primarily from the manufacturing imperfections and the inhomogeneity of silicon wafers. It can not be affected by the environment. However, two things need to be considered for these methods based on SPN. Firstly, the quality of SPN extracted from image depends on the image content. And the second one is that the detection performance could be decreased with the reduction of image size. 

The convolutional neural network(CNN) has recently achieved better performance than traditional schemes in digital image forensics \cite{16,17,18,19}. There are two common characters for these algorithms. Firstly, considering the different tasks between computer vision and image forensics, the researches focusing on image forensics usually add preprocessing operations into convolutional nerual network architecture, which can amplify the inter-class difference and reduce the impact of the image content. For example, median filter, Laplacian filter, and high-pass filter are applied in median filtering forensics, recapture forensics, and source camera identification, respectively. Secondly, according to the reports in these works, the convolutional nerual networks is suit for dealing with small-size images.  

In this paper, we propose a content-adaptive fusion network to achieve the source camera identification for small-size images. Firstly, we choose a convolutional layer in preprocessing stage. Because that the special filtering operation in preprocessing stage is for amplifying the interclass difference and reduce the impact of the image content. However, some useful informations inside the images could also be lost. Adding the convolutional layer in perprocessing stage can makes the convolutional neural networks learn content-adaptive convolutional kernels from input data. What's more, in order to capture more comprehensive information, three adaptive-content convolutional neural networks, CA3-CNN, CA5-CNN, CA7-CNN, are paralleled together to construct the content-adaptive fusion network(CAF-CNN). The difference between three CA-CNNs is the convolutional kernel size in the preprocession. The effectiveness of proposed method is vaildated in the experiments. The detection performances of proposed method on four kinds of cases are discussed: camera brand idenfication, camera model idenfication, camera device identification, and source camera identification that fusing different brand, model, and device. The experimental results show that the proposed method is practicable and satisfactory.

The rest of this paper is organized as follows. Section 2 describes the related work; Section 3 presents the details of the algorithm proposed in this work; Section 4 includes the experimental results; conclusions are given in section 5.

\section{RELATED WORKS}
\label{sec:format}
\begin{figure*}[!tp]
	\centering
	\includegraphics[width=14cm,height=6cm]{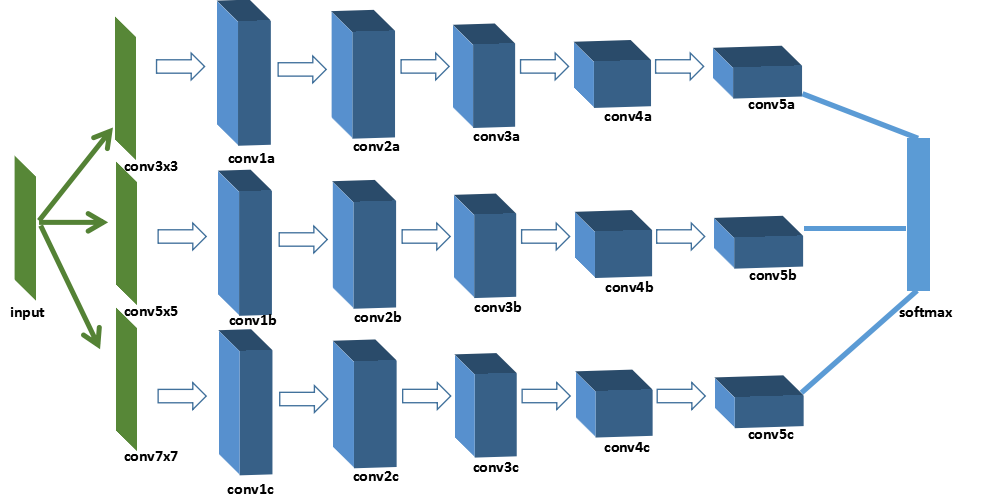}
	\caption{the architecture of content-adaptive fusion network}
\end{figure*}

A preprocessing layer is added to the CNN architecture in the work\cite{19}. And two types of residuals in preprocessing stage are evaluated: high-pass filtering residual and the residual noise extracted by subtracting the denoised version of the image from the image itself. The two residuals are shown in the following formulars, respectively.
\begin{equation}\label{key}
R_{1}=I-WF(I)
\end{equation}

\begin{equation}\label{key}
R_{2}=I*\dfrac{1}{12}
\begin{bmatrix}
-1 & 2 & -2 & 2 & -1 \\ 2 & -6 & 8 & -6 & 2 \\-2 & 8 & -12 & 8 & -2 \\2 & -6 & 8 & -6 & 2 \\-1 & 2 & -2 & 2 & -1
\end{bmatrix}
\end{equation}
Where I repensent the input image; WF(I) is the denoised image as descripted in Lukas' work \cite{20}; * mean the convoluting operation. As reported in this work, adding high-pass filter in the pre-procession into the CNN architecture has the better detection performance.

In our previous work\cite{17}, we proposed the LCNN to detect recaptured images. The LCNN has better detection performance than the algorithms based on handcrafted features. The architecture of the LCNN is shown in Fig 1. There has a preprocessing layer, five convolutional layers and a softmax layer. Laplician filtering operation is put into preprocessing layer. The convolutional layer contains four operations: convolution, Batch-Normalization, ReLu, and average pooling. The numbers of feature maps in five convolutional layers are 8, 16, 32, 64, and 128, respectively. In order to avoid overfitting, we applied  global average pooling to the last pooling layer and directly fed the output of global average pooling into softmax layer. What's more, the Batch-Normalization layer is used. It has been proved that it is an effective mode to accelerate convergence. Owing to the generalization of convolutional nerual networks, we also make use of the LCNN architecture in this work.

\section{PROPOSED ALGORITHM}
\label{sec:pagestyle}
 According to the image forensics algorithms based on CNN, in order to magnify the inter-class differences, the preprocessing layer is added into the CNN architecture. And the special filter is applied in the preprocessing stage. For example, Laplician filter and high-pass filter are used for recapture forensics, camera model identification, respectively. The special filter is good for amplifying the inter-class differences, but it could also be a double-edged sword. Because it maybe drop some useful information. Considering that the convolutional neural network can self-learn better feature reprensentations from input data, we replace the special filter with the content-adaptive filter, which can self-learn the convolutional kernels according to the input data. What's more, the combination of different convolutional kernel sizes in preprocessing stage maybe capture the various information. So we build a content-adaptive fusion network to learn more comprehensive features by paralleling three CA-CNNs together. 
\subsection{CONTEND-ADPTIVE CONVOLUTIONAL NERUAL NETWORKS}
The filtering kernel should not be same for different input data. For example, for source camera identification, the traditional schemes calculate the correlation between SPN and the reference SPN. It is important for detection performance to extract the high-quality SPN. And it has been proved that the SPN is related to the image content. The different image contents will be dealed with different ways. Therefore, it is maybe not the best way to preprocess input data using high-pass filter for source camera identification based on CNN. Considering that the convolutional neural network can self-learn better feature reprensentations from input data, in this work, we design a content-adaptive convolutional neural networks(CA-CNN) by replacing the special filter of LCNN with a convolutional layer, as is shown in the following formula:
\begin{equation}\label{key}
R=I*W_{pq}
\end{equation}
Where, W means the values of the convolutional kernel and it can be learned from the input data using mini-batch gradient descent. The architecture of CA-CNN is shown in Fig 1. We test it in three dataset. In Fig. 3, there are four kinds of convolutional kernels. As we pointed out, the convolutional kernels in the preprocessing stage for different input data should not be same. The high-pass filter is shown in Fig. 3(a) and other filters are learned from three datasets. It is obvious that content-adptive convolutional neural networks can self-learn filters from the different input data. 
\begin{figure}
	\begin{minipage}[t]{0.45\linewidth}
	\centering
	\includegraphics[height=2.5cm,width=2.5cm]{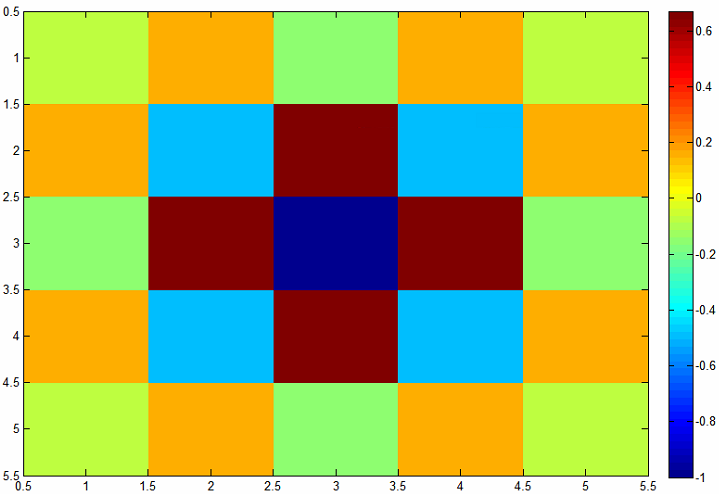}
	\centerline{(a)}
\end{minipage}
\hfil
\begin{minipage}[t]{0.5\linewidth}
	\centering
	\includegraphics[height=2.5cm,width=2.5cm]{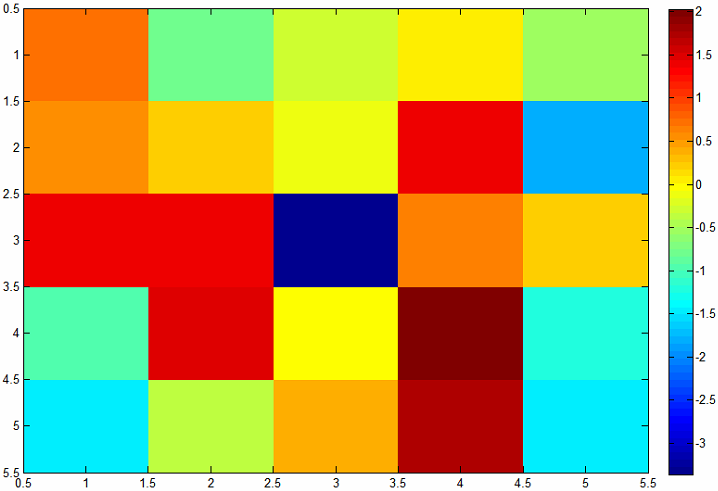}
	\centerline{(b)}
\end{minipage}
\begin{minipage}[t]{0.45\linewidth}
	\centering
	\includegraphics[height=2.5cm,width=2.5cm]{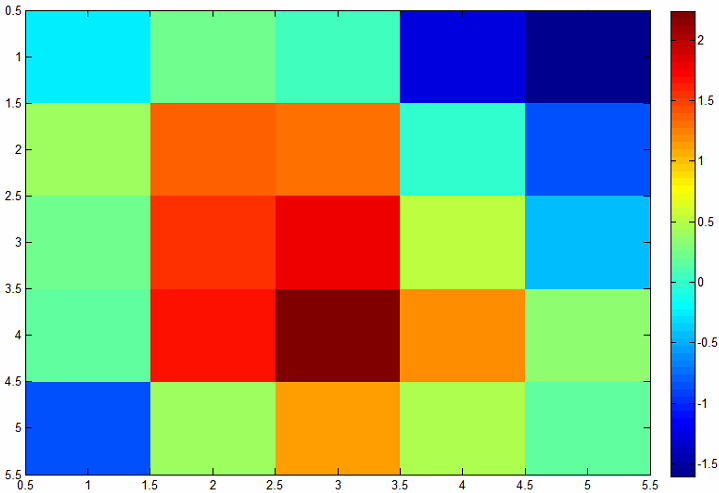}
	\centerline{(c)}
\end{minipage}
\hfill
\begin{minipage}[t]{0.5\linewidth}
	\centering
	\includegraphics[height=2.5cm,width=2.5cm]{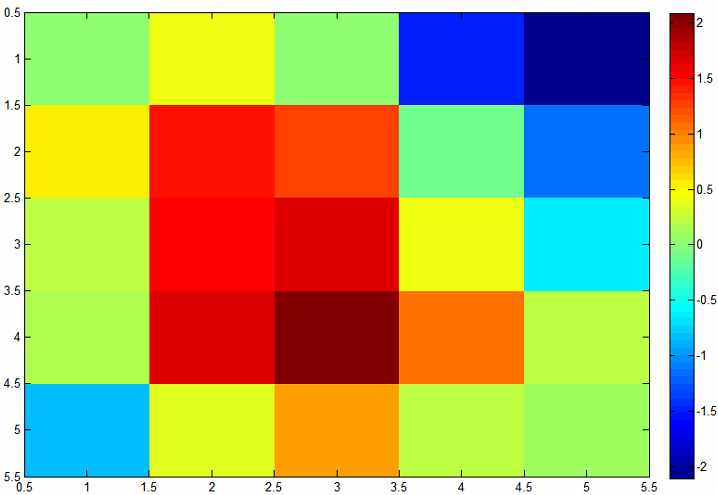}
	\centerline{(d)}
\end{minipage}
\caption{filter visualization. high-pass filter (a), content-adaptive filter (b) (c) (d) learned from Part 4.1 and 4.2, respectively.}
\end{figure}

\label{sssec:subsubhead}
\subsection{CONTEND-ADPTIVE FUSION NETWORK}
\label{sssec:subsubhead}
In order to capture the more comprehensive features, we parallel three content-adptive convolutional neural networks, as is shown in Fig. 2. The input image is processed in pre-processing layer by three three kinds of convolutional kernel size: 3x3, 5x5, 7x7, respectively. Then five basic units are used. Basic unit includes four part: convolution, Batch-Normalization, ReLu, and average pooling operation. The numbers of feature maps in five basic units are 8, 16, 32, 64, and 128, respectively. In order to reduce the number of model paramaters, global average pooling operation is used in conv5a, conv5b, and conv5c layers. The output of global average pooling are put together and then fed into a softmax layer. The calculation of softmax layer is as follwing:
\begin{equation}\label{key}
f(y_{i})=-log(\dfrac{e^{y_{i}}}{\sum_{i=1}^{n}e^{y_{i}}})
\end{equation}
\begin{equation}\label{key}
y_{i}=\sum_{k=1}^{384} W_{k}^{i}*X_{k}+B_{i}
\end{equation}
where, i represents the class label, n is the number of the classes; X and k mean the output of global average pooling and its number, respectively; W and B are the weights and biases, which will be learned using mini-batch gradient descent.

\section{EXPERIMENTS}
\label{sec:typestyle}
In order to validate the proposed algorithm, we conduct a set of experiments on the Dresden Database that provide more than 16000 images took by 74 camera devices. A new dataset is constructed by cropping the image into 64x64 patches. In this work, we choose 13 camera devices and the list of camera devices are given in Tabel 1. There are nine camera brands, thirteen camera devices. The dataset is splited by assigning 4/6 of the images to a training set, 1/6 to a validation set, and 1/6 to a test set, respectively. The detection accuracies are averaged over 3 random experiments. The learning rate is initialized to 0.01, and scheduled to decrease 10\% for every 10000 iterations. The max iteration is set to 500000 and the momentum is fixed to 0.9.

\begin{table}
	\centering
	\caption{The list of camera devices used}
	\begin{tabular}{lccc}
		\hline
		ID & Camera Devices & Original Resolution\\ \hline
		1 & Kodak\_M1063\_0 & 3664x2748 \\ \hline
		2 & Pentax\_OptioA40\_0 & 4000x3000\\ \hline
		3 & Nikon\_CoolPixS710\_1 & 4352x3264  \\ \hline
		4 & Sony\_DSC-H50\_0 & 3456X2592\\ \hline
		5 & Olympus\_mju\_1050SW\_2 & 3648x2736\\ \hline
		6 & Panasonic\_DMC-FZ50\_1 & 3648x2736\\ \hline
		7 & Agfa\_Sensor530s\_0 & 2560x1920  \\ \hline
		8 & Ricoh\_GX100\_0 & 3648x2736\\ \hline
		9 & Samsung\_NV15\_0 & 3648x2736 \\ \hline
		10 & Sony\_DSC-W170\_0 & 3648x2736\\ \hline
		11 & Sony\_DSC-T77\_0 & 3648x2736 \\ \hline
		12 & Sony\_DSC-T77\_1 & 3648x2736\\ \hline
		13 & Sony\_DSC-T77\_2 & 3648x2736\\ \hline
		
	\end{tabular}

\end{table}

\subsection{CAMERA BRAND IDENTIFICATION}
\label{ssec:subhead}
In the first experiment, in order to assess the performance of the algorithm in the case of camera brand identification, nine camera devices from ID 1 to ID 9 are selected. The detection accuracies are shown in Table 2. HP-CNN represent the architecture with high-pass filter in pre-processing layer. CA3-CNN, CA5-CNN, and CA7-CNN mean the content-adaptive convolutional neural networks with 3x3, 5x5, 7x7 convolutional kernel in pre-processing layer, respectively. CAF-CNN is the content-adaptive fusion network. According to the results, it can confirm that content-adaptive filtering operation is a good way for CNN framework. What's more, the different kernel size in preprocessing stage can capture the more comprehensive features. So CAF-CNN attain the best detection performance.

\begin{table}[]
	\centering
	\caption{The detection accuracy for camera brand identification. The best results are highlighten in bold.}
	\label{my-label}
	\begin{tabular}{|c|p{1.1cm}|p{1.1cm}|p{1.1cm}|p{1.1cm}|p{1.1cm}|}
		\hline
		& HP-CNN & CA3-CNN & CA5-CNN & CA7-CNN & CAF-CNN \\ \hline
		1 & 85.83\% & 93.63\% & 91.26\% & 92.01\% & \textbf{96.37\%} \\ \hline
		2 & 93.83\% & 90.55\% & 93.61\% & 96.16\% & \textbf{97.76\%} \\ \hline
		3 & 75.23\% & 86.06\% & 91.64\% & 91.72\% & \textbf{96.39\%} \\ \hline
		4 & 83.09\% & 87.52\% & 92.00\% & 90.39\% & \textbf{93.86\%} \\ \hline
		5 & 78.38\% & 78.27\% & 80.02\% & 82.99\% & \textbf{89.56\%} \\ \hline
		6 & 80.58\% & 87.35\% & 88.50\% & 92.88\% & \textbf{94.64\%} \\ \hline
		7 & 86.29\% & 91.76\% & 93.66\% & 92.59\% & \textbf{94.65\%} \\ \hline
		8 & 81.31\% & 91.83\% & 94.20\% & 90.82\% & \textbf{94.64\%} \\ \hline
		9 & 70.03\% & 82.55\% & 86.06\% & 86.58\% & \textbf{89.70\%} \\ \hline
		AVE & 81.62\% & 87.72\% & 90.11\% & 90.68\% & \textbf{94.17\%} \\ \hline
	\end{tabular}
\end{table}

\subsection{CAMERA MODEL, DEVICE IDENTIFICATION}
\label{ssec:subhead}
In the second experiment, the performances of proposed method for camera model and device identification are evaluated. For the case of camera model identification, three camera models are selected: Sony\_DSC-H50, Sony\_DSC\-W170, and Sony\_DSC-T77. For the case of camera device identification, three camera devices from the same model are chose: Sony\_DSC-T77\_0, Sony\_DSC-T77\_1, and Sony\_DSC-T77\_2. In stead of re-training a new CAF-CNN model, we finetune the model trained in the first experiment.

For camera model identification, in order to keep balance between Sony\_DSC-H50\_0 and the others, we randomly select 409174 images from the images took by Sony\_DSC-H50\_0. The detection accuracy is above 84.7\%. For camera device identification, three devices of Sony\_DSC-T77 is used. There is no doubt that camera device identification is a hard task. The average detection accuracy of the proposed method in this case is 70.19\%.

In order to further estimate the feasibility of the algorithm, we test it in the mixing dataset, including different camera brands, same camera brand but different camera models, and same camera model but different camera devices. The cameras used in this stage are Sony\_DSC-T77\_0, Sony\_DSC-T77\_1, Sony\_DSC-H50\_0,    Olympus\_mju\_1050SW\_2, Panasonic\_DMC-FZ50\_1, Agfa\_Sensor530s\_0, Ricoh\_GX100\_0, Samsung\_NV15\_0, Kodak\_M1063\_0. The detection performance is near 87\%, which demonstrate that it is practicable and satisfactory to identify source camera for small-size images.

\section{CONCLUSIONS}
\label{sec:majhead}

In this paper, we propose a content-adaptive fusion network for small-size images to achieve the source camera identification. In order to learn better feature representation from the input data, content-adaptive convolutional neural networks are constructed. we add one convolutional layer into pre-processing stage and the parameters of convolutional kernel need to be learned from the input data. What's more, the content-adaptive fusion network is built by paralleling three adaptive-content convolutional neural networks to capture more comprehensive information. The experimental results show that the proposed algorithm can identify the camera brand, camera model, camera device and source camera for small-size images. We believe that fusing the other effective CNN, such as GoogleNet\cite{21} would still work, and the architecture of CAF-CNN could be applied to other image forensics scenarios.



\bibliographystyle{unsrt}
\bibliography{ICIP-Source}

\end{document}